\documentclass[a4paper, 10pt, conference]{ieeeconf}      
\usepackage{FG2026}

\FGfinalcopy 

\IEEEoverridecommandlockouts                              
\usepackage{amsmath} 
\usepackage{amssymb}  
\usepackage{graphicx}
\usepackage{comment}
\usepackage{hyperref}
\let\labelindent\relax
\usepackage{enumitem}
\usepackage{cleveref}

\usepackage{booktabs}
\usepackage{threeparttable}
\usepackage{xcolor}
\usepackage{pifont}
\usepackage[table]{xcolor}
\usepackage{url}

\usepackage{amssymb}
\usepackage{pifont}
\definecolor{ForestGreen}{RGB}{34,139,34}
\definecolor{lightblue}{RGB}{230,242,255}
\newcommand{\cmark}{\textcolor{ForestGreen}{\ding{51}}}
\newcommand{\xmark}{\textcolor{red}{\ding{55}}}

\def\FGPaperID{420} 

\usepackage{booktabs}       
\usepackage{tabularx}       
\usepackage{multirow}       
\usepackage{threeparttable} 

\title{\LARGE \bf
Geometry-Conditioned Diffusion for Occlusion-Robust \\ In-Bed Pose Estimation
}

\author{\parbox{16cm}{\centering
    {\large Navid Aslankhani Khameneh$^{1}$, Marco Carletti$^2$, and Cigdem Beyan$^1$}\\
    {\normalsize
    $^1$Department of Computer Science, University of Verona, Verona, Italy \\
    $^2$EVS - Embedded Vision Systems Srl, Verona, Italy}}
}

\usepackage{fancyhdr}
\begin{document}

\ifFGfinal
\thispagestyle{empty}
\pagestyle{empty}
\else
\author{Anonymous FG2026 submission\\ Paper ID \FGPaperID \\}
\pagestyle{plain}
\fi
\maketitle
\thispagestyle{fancy}
\renewcommand{\headrulewidth}{0pt}
\fancyhf{}
\fancyhead[C]{}


\begin{abstract}
Robust in-bed human pose estimation under blanket occlusion remains challenging due to the scarcity of reliable labeled training data for heavily covered poses. Existing approaches rely on multi-modal sensing or image-to-image translation frameworks that remain conditioned on visible source imagery, limiting scalability and pose diversity. In this work, we reformulate occlusion-aware augmentation as a geometry-conditioned generative modeling task. We conduct a systematic comparison of deterministic masking, unpaired translation, paired diffusion-based translation, and a proposed pose-conditioned Latent Diffusion Model (Pose-LDM). Unlike image-guided methods, Pose-LDM synthesizes blanket-covered images directly from skeletal keypoints, eliminating dependence on paired supervision and pixel-level source-image conditioning while enabling generation from arbitrary pose inputs. All augmentation strategies are evaluated through their impact on downstream pose estimation under a fixed backbone. 
Pose-LDM achieves the highest strict localization accuracy under severe occlusion while maintaining overall detection performance comparable to paired diffusion models, approaching the performance of fully supervised training.
These results demonstrate that geometry-conditioned diffusion provides an effective and supervision-efficient pathway toward occlusion-robust in-bed pose estimation without modifying the sensing pipeline. The code is available at: \url{github.com/navidTerraNova/GeoDiffPose}. 
\end{abstract}

\section{Introduction}
\label{sec:intro}

Robust in-bed human pose estimation is a key component of non-invasive patient monitoring systems, enabling applications such as sleep analysis, fall prevention, and long-term mobility assessment~\cite{liu2020slp,yin2020}. While modern RGB-based pose estimators perform remarkably well under standard conditions~\cite{maji2022yolopose,cao2019openpose}, their performance typically deteriorates in realistic hospital and home environments where patients are frequently covered by blankets. Severe occlusion fundamentally alters the visual evidence available to the model: instead of directly observing limbs and joints, the system must infer body configuration indirectly from subtle surface deformations of fabric.

This challenge is not merely one of visual complexity but of supervision. Under heavy blanket occlusion, annotators cannot reliably label joint positions that are not visible, making the collection of accurate large-scale ``covered'' datasets inherently difficult~\cite{cao2021,afham2022}. As a result, training robust models for this setting faces a structural data bottleneck.

Existing approaches address this limitation in two main ways. One direction relies on multi-modal sensing, such as depth, thermal, or pressure maps, to mitigate the visibility issue by leveraging alternative sensing cues~\cite{dayarathna2022,davoodnia2021,yin2020,obeidavi2022pose}.
While effective, these systems require specialized hardware, strict calibration, and controlled acquisition environments, limiting scalability in real-world home-care settings. Another direction leverages generative image-to-image translation models to synthesize blanket occlusions from uncovered infrared images~\cite{afham2022}. These methods reduce the need for manual labeling by transferring existing ground-truth annotations to synthetically covered samples. However, image-guided translation remains fundamentally constrained by its conditioning signal: it requires a visible source image and can only generate occlusions aligned with the poses already present in the dataset.

This observation motivates an important question:
\begin{center}
\emph{Can we synthesize realistic blanket occlusions directly from skeletal geometry, without conditioning on a visible source image?}
\end{center}

If generation is conditioned solely on skeletal geometry rather than pixel-level input, several potential advantages arise. First, synthesis becomes decoupled from the availability of visible source images. Second, novel pose configurations can be generated at inference time by modifying the input keypoints. Third, such a formulation reduces reliance on identifiable appearance information, as generation is driven by abstract skeletal representations rather than raw patient imagery.

To systematically analyze the role of conditioning signals in occlusion synthesis, we conduct a comparative study spanning deterministic masking, image-guided translation, and geometry-conditioned generation. Importantly, all augmentation strategies are evaluated based on their impact on downstream in-bed pose estimation performance using the same pose estimation backbone, thereby isolating the contribution of the generative model. We begin with a heuristic baseline to quantify the minimal gains achievable through simple blanket-like masking. We then evaluate image-guided generative models in unpaired (CycleGAN)~\cite{zhu2017} and paired (Brownian Bridge Diffusion Model, BBDM)~\cite{li2023} settings to assess the benefits of learned stochastic texture synthesis and structural alignment. Finally, we introduce a pose-conditioned Latent Diffusion Model (LDM)~\cite{rombach2022high} that synthesizes covered images directly from the skeletal keypoint coordinates.

In contrast to image-to-image translation methods, i.e., CycleGAN and BBDM, which generate covered samples by transforming an existing uncovered image at the pixel level, our pose-conditioned framework (Pose-LDM) eliminates source-image conditioning entirely. Instead of learning a mapping between the uncovered and covered image domains, the generative process operates in latent space and injects skeletal information through learned keypoint embeddings. This enables the model to synthesize blanket occlusions directly from pose geometry while preserving the underlying body configuration required for valid supervision in downstream pose estimation. Consequently, data augmentation is reformulated from a domain translation problem into a geometry-conditioned synthesis problem, allowing improvements in occlusion robustness to be attributed directly to the conditioning paradigm rather than image-level alignment.

Our experiments in the SLP dataset~\cite{liu2020slp} demonstrate three key findings for the estimation of downstream poses under blanket occlusion.
First, Pose-LDM achieves the highest joint localization accuracy under severe occlusion, outperforming heuristic masking, unpaired translation, and paired diffusion-based translation.
Second, although the paired BBDM obtains slightly higher overall detection scores, Pose-LDM delivers comparable performance without requiring aligned uncovered-covered image pairs or pixel-level source conditioning.
Third, we show that the way pose information is integrated into the diffusion model impacts performance, indicating that conditioning design plays a critical role in geometry-preserving occlusion synthesis.

These results demonstrate that blanket synthesis conditioned solely on skeletal geometry can generate synthetic covered images that effectively support occlusion-robust pose training. More broadly, they show that geometry-only conditioning in diffusion models is sufficient to achieve competitive performance without requiring paired images or pixel-level source conditioning.

In summary, this paper reframes occlusion-robust in-bed pose estimation not as a multi-modal sensing problem nor purely as an image translation problem, but as a geometry-conditioned generative modeling task. By removing dependence on visible source images and demonstrating competitive performance with pose-only conditioning, we provide a practical pathway for synthetic data generation in heavily occluded scenarios. While motivated by in-bed monitoring, the proposed geometry-conditioned diffusion framework is broadly applicable to gesture and body analysis tasks where occlusion-aware synthetic data is required.

\noindent Our main contributions are summarized as follows:
\begin{itemize}[leftmargin=*, nosep]
    \item We reformulate occlusion-aware in-bed pose augmentation as a geometry-conditioned generative modeling problem, removing dependence on visible source images.
    
    \item We introduce a pose-conditioned Latent Diffusion Model that synthesizes blanket occlusions directly from raw keypoint coordinates while preserving pose supervision for downstream training.
    
    \item We demonstrate that pose-only diffusion achieves performance competitive with strictly paired diffusion models, enabling data augmentation without requiring paired uncovered–covered images.
    
    \item We provide a systematic comparison across heuristic, GAN-based, paired diffusion, and pose-conditioned augmentation strategies, and show that conditioning design within the diffusion model impacts downstream pose estimation performance.
\end{itemize}

\section{Related Work}
\label{sec:relatedwork}

Estimating human pose under blanket occlusion has been studied through alternative sensing modalities and multi-modal fusion strategies. A substantial body of work addresses the visibility limitations of RGB imagery by incorporating privacy-preserving or non-RGB sensing modalities such as LWIR, depth, and pressure maps. For example, Dayarathna et al.~\cite{dayarathna2022} propose a multi-sensor fusion framework that integrates LWIR, depth, and pressure signals, followed by a conditional GAN to reconstruct visible images prior to pose estimation. Similarly, Yin et al.~\cite{yin2020} combine aligned depth, LWIR, pressure, and RGB inputs within a coarse-to-fine pyramid architecture to recover 3D Skinned Multi-Person Linear (SMPL) body shapes~\cite{loper2023smpl} under blankets, incorporating an attention-based reconstruction module to mitigate occlusion effects. While these methods demonstrate strong performance, they rely on synchronized multi-sensor acquisition and tightly aligned training data, increasing hardware complexity and limiting scalability beyond controlled laboratory or clinical environments.

Rather than fusing modalities at inference, other approaches aim to reconstruct missing modalities to enable single-modality deployment. Cao et al.~\cite{cao2021} introduce a Multi-Modal Conditional Variational Autoencoder that reconstructs RGB feature representations from LWIR inputs, allowing pose estimation when visible imagery is unavailable. Although effective for handling absent modalities, such frameworks remain dependent on paired multi-modal training data and operate by completing missing modality representations rather than addressing the limited availability of reliable labeled data under heavy blanket occlusion.

Other works replace RGB entirely with alternative sensing modalities. Davoodnia et al.~\cite{davoodnia2021} propose PolishNetU, a domain adaptation network that transforms pressure maps into image-like representations compatible with pre-trained pose estimators such as OpenPose~\cite{cao2019openpose}. Similarly, Obeidavi et al.~\cite{obeidavi2022pose} employ a Multi-Scale Stacked Hourglass network~\cite{newell2016stacked} to estimate keypoints directly from LWIR imagery. These approaches avoid visible-light limitations and demonstrate robustness to darkness and coverage conditions. However, they require specialized sensing hardware and focus on modality substitution or domain adaptation rather than enhancing RGB-based pose estimators under occlusion.

Beyond sensor-driven solutions, several works seek to improve occlusion robustness through learning-based strategies within a single modality. Afham et al.~\cite{afham2022} propose CycAug, an unpaired image-to-image translation framework based on CycleGAN~\cite{zhu2017}, to synthesize labeled covered LWIR images from uncovered LWIR samples. Their method combines translation-based augmentation with self-supervised knowledge distillation to reduce cross-domain discrepancies between uncovered and covered thermal imagery. While effective in mitigating annotation scarcity, the approach remains conditioned on visible source images and operates as a domain adaptation mechanism, thereby restricting synthesized samples to pose configurations present in the uncovered dataset.

Instead of image translation, Bigalke et al.~\cite{bigalke2022} address occlusion through anatomy-guided domain adaptation for 3D in-bed pose estimation from point clouds. Their framework adapts a source-trained model to a shifted target domain by enforcing anatomical plausibility constraints and filtering pseudo labels within a Mean Teacher paradigm~\cite{tarvainen2017mean}. Although effective in reducing domain gaps between uncovered and covered settings, this strategy modifies model parameters to fit the target distribution rather than generating new training samples.

Collectively, existing methods mitigate blanket occlusion through sensor substitution, cross-modal reconstruction, image translation, or model-level adaptation. In contrast, we focus exclusively on the RGB domain and address the scarcity of reliable labeled training data under blanket occlusion without altering the sensing pipeline. While thermal, depth, and pressure-based systems provide robustness under occlusion, they require dedicated hardware and calibration procedures that hinder scalable deployment, particularly in home-care environments where RGB cameras remain the most accessible and economically viable modality. By reformulating occlusion robustness as a geometry-conditioned generative modeling problem, we synthesize blanket-covered samples directly from skeletal keypoints, decoupling geometry from appearance and enabling the generation of previously unseen pose configurations. This geometry-driven augmentation is evaluated through its impact on downstream pose estimation accuracy, allowing improvements in occlusion robustness to be attributed to the conditioning paradigm rather than modifications to the pose estimator itself.

\section{Methodology}
\label{sec:method}

\subsection{Problem Formulation}
Let $\mathcal{U}$ denote the domain of uncovered RGB images and $\mathcal{C}$ the domain of blanket-covered images. For each uncovered image $u \in \mathcal{U}$, we assume access to reliable ground-truth pose keypoints $p \in \mathbb{R}^{K \times 2}$, where $K$ denotes the number of body joints. In contrast, accurate annotations for images in $\mathcal{C}$ are difficult to obtain due to severe occlusion.

Our objective is to synthesize realistic covered samples $\hat{c} \in \mathcal{C}$ while preserving the underlying body geometry defined by $p$, such that the original supervision can be transferred without manual re-annotation. For image-guided methods, the generator learns a mapping $f_\theta : \mathcal{U} \rightarrow \mathcal{C}$, whereas for pose-conditioned generation, the mapping is defined as $f_\theta : \mathbb{R}^{K \times 2} \rightarrow \mathcal{C}$. In both cases, the augmentation is designed to preserve the pose geometry so that the ground-truth keypoints associated with the uncovered images remain valid for the synthesized samples.

\subsection{Generative Augmentation Strategies}
The core objective of our approach is to mitigate the performance degradation of RGB-based pose estimators under blanket occlusion without relying on manually labeled covered data. To this end, we investigate data augmentation strategies of progressively increasing modeling capacity, moving from a heuristic baseline to image-guided translation models, and ultimately to a pose-conditioned diffusion framework.

Across all strategies, the augmentation process is designed to preserve the underlying body geometry during synthetic occlusion.
This enables direct transfer of ground-truth keypoint annotations from uncovered images to synthesized covered samples, allowing supervised training without additional manual labeling.

\subsubsection{Heuristic Baseline}

\begin{figure}[tb!]
    \centering
    \includegraphics[width=\linewidth]{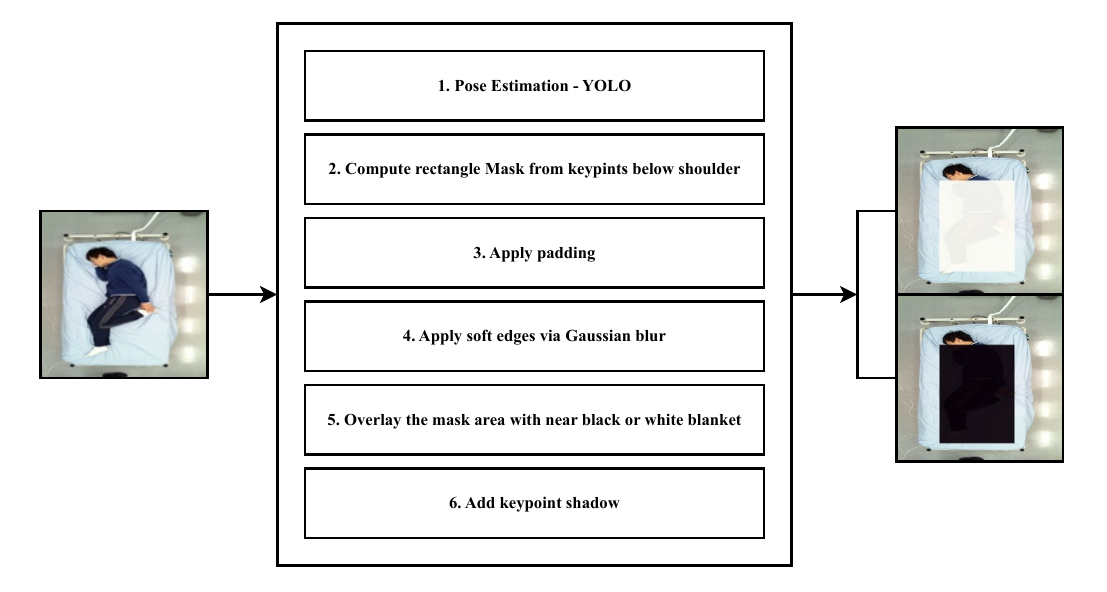}
    \caption{Overview of the heuristic blanket augmentation baseline. Lower-body keypoints predicted by a pre-trained pose estimator define a padded rectangular mask, which is smoothed using Gaussian filtering and blended with a near-uniform blanket color (black or white). Local intensity attenuation around selected joints simulates minimal contact shadows.}
    \label{fig:heuristic}
\end{figure}

As a lower-bound baseline, we simulate blanket occlusion using deterministic image processing operations (see Fig.~\ref{fig:heuristic}). This baseline establishes a minimal reference performance without learned generative modeling.
The augmentation pipeline begins by extracting pose keypoints from the uncovered image using a pre-trained pose estimator, which serves as a geometric anchor for mask construction. We emphasize that these predicted keypoints are used solely to define the occlusion region, while ground-truth annotations from the dataset used remain unchanged and are used for supervision during pose estimator training. We identify the subset of keypoints $\mathcal{K}_{sub}$ located anatomically below the shoulders, approximating the body region typically covered by a blanket.

We define a padded bounding region operator $\mathcal{R}(\cdot)$ as:
\begin{equation}
\mathcal{R}(\mathcal{K}_{sub}) =
[x_{\min}-\delta_x,\; x_{\max}+\delta_x] \times
[y_{\min}-\delta_y,\; y_{\max}+\delta_y],
\end{equation}
where $x_{\min} = \min_i x_i$, $x_{\max} = \max_i x_i$, 
$y_{\min} = \min_i y_i$, and $y_{\max} = \max_i y_i$ 
over all $(x_i, y_i) \in \mathcal{K}_{sub}$, 
and $\delta_x, \delta_y \sim \mathcal{U}(a,b)$ denote independent random padding variables sampled from a uniform distribution over $[a,b]$, where $a$ and $b$ define the minimum and maximum padding (in pixels).

The raw binary mask $M_{raw}$ is then defined as:
\begin{equation}
M_{raw}(x,y) =
\begin{cases}
1 & \text{if } (x,y) \in \mathcal{R}(\mathcal{K}_{sub}) \\
0 & \text{otherwise}
\end{cases}
\end{equation}
To mimic the soft transition of fabric boundaries, the mask is convolved with a Gaussian kernel $G_\sigma$ with standard deviation $\sigma$:
$M_{soft} = M_{raw} * G_\sigma$.

The augmented image $I_{aug}$ is then obtained by blending the original image $I$ with a near-uniform blanket color $C_{blanket}$: $I_{aug} = (1 - M_{soft}) \odot I + M_{soft} \odot C_{blanket}.$

Finally, to introduce minimal depth cues, we locally attenuate intensity around selected joint locations, simulating contact shadows cast by the body beneath the blanket.

Although computationally efficient, this approach lacks realistic texture variation, fold structure, and physically plausible shading. It therefore serves as a controlled lower bound for evaluating learned generative models.
\\

\subsubsection{Image-Guided Synthesis (Uncovered $\rightarrow$ Covered)}
We next consider image-guided translation models that learn a mapping from the uncovered domain $\mathcal{U}$ to the covered domain $\mathcal{C}$ while preserving spatial structure.

\begin{figure}[t]
    \centering
    \includegraphics[width=\linewidth]{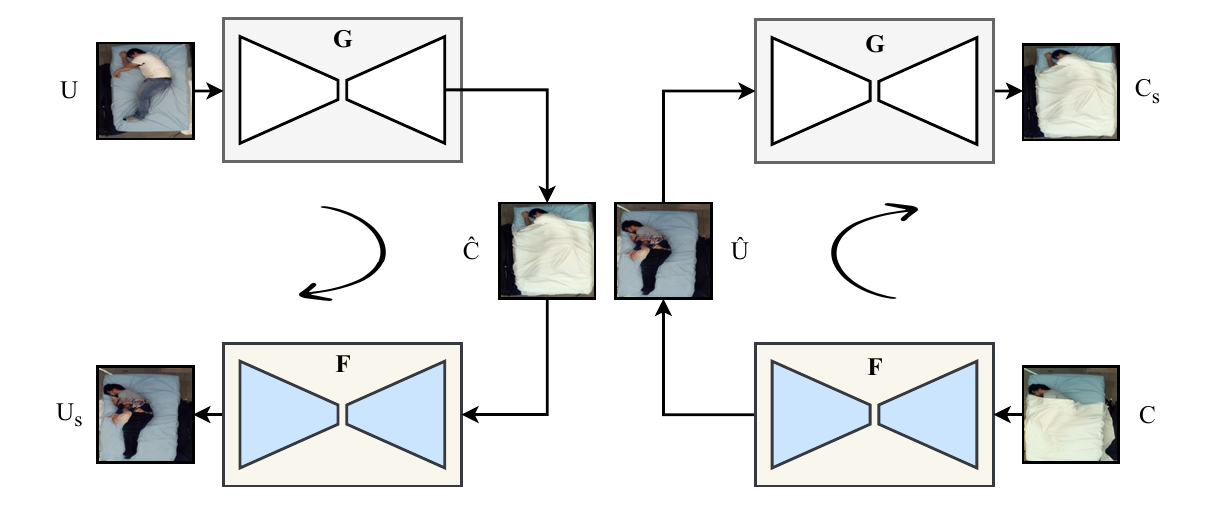}
    \caption{
    CycleGAN-based unpaired image-to-image translation for blanket synthesis. Two generators are learned: $G:\mathcal{U}\rightarrow\mathcal{C}$ maps uncovered images $U$ to the covered domain, and $F:\mathcal{C}\rightarrow\mathcal{U}$ maps covered images $C$ back to the uncovered domain. Cycle-consistency enforces $F(G(u)) \approx u$ and $G(F(c)) \approx c$. During inference, only $G$ is used to generate synthetic covered samples $C_s = G(U)$ from uncovered inputs.  
    }
    \label{fig:cyclegan}
\end{figure}

\paragraph{Unpaired Translation (CycleGAN)}
We employ CycleGAN~\cite{zhu2017} as a canonical unpaired image-to-image translation baseline to represent pixel-level domain translation without paired supervision (see Fig.~\ref{fig:cyclegan}). CycleGAN has been widely adopted in diverse translation settings, including recent extensions and analyses~\cite{torbunov2023uvcgan,dou2020asymmetric,nyamathulla2024analysis,pang2021image}. It learns two generators, $G:\mathcal{U}\rightarrow\mathcal{C}$ and $F:\mathcal{C}\rightarrow\mathcal{U}$, without requiring paired training samples.

To preserve structural consistency, the model enforces a cycle-consistency loss:
\begin{equation}
\begin{split}
\mathcal{L}_{cyc}(G, F) &=
\mathbb{E}_{u \sim p_{\mathcal{U}}}
\|F(G(u)) - u\|_1 \\
&\quad +
\mathbb{E}_{c \sim p_{\mathcal{C}}}
\|G(F(c)) - c\|_1.
\end{split}
\end{equation}

The full objective combines adversarial losses with the cycle-consistency constraint:
\begin{equation}
\begin{split}
\mathcal{L}(G, F, D_U, D_C) &=
\mathcal{L}_{GAN}(G, D_C) \\
&\quad + \mathcal{L}_{GAN}(F, D_U) \\
&\quad + \lambda \mathcal{L}_{cyc}(G, F).
\end{split}
\end{equation}
where $\mathcal{L}_{GAN}$ denotes the standard adversarial loss formulation as in~\cite{zhu2017}, and $\lambda$ controls the relative strength of the cycle-consistency constraint.

During inference, only the generator $G$ is used to synthesize covered samples $C_s = G(U)$.

Although CycleGAN can generate plausible blanket textures and global appearance changes, the synthesis remains conditioned on the visible source image. Consequently, generated occlusions are limited to pose configurations already present in the uncovered dataset.

\begin{figure}[tb!]
    \centering
    \includegraphics[width=\linewidth]{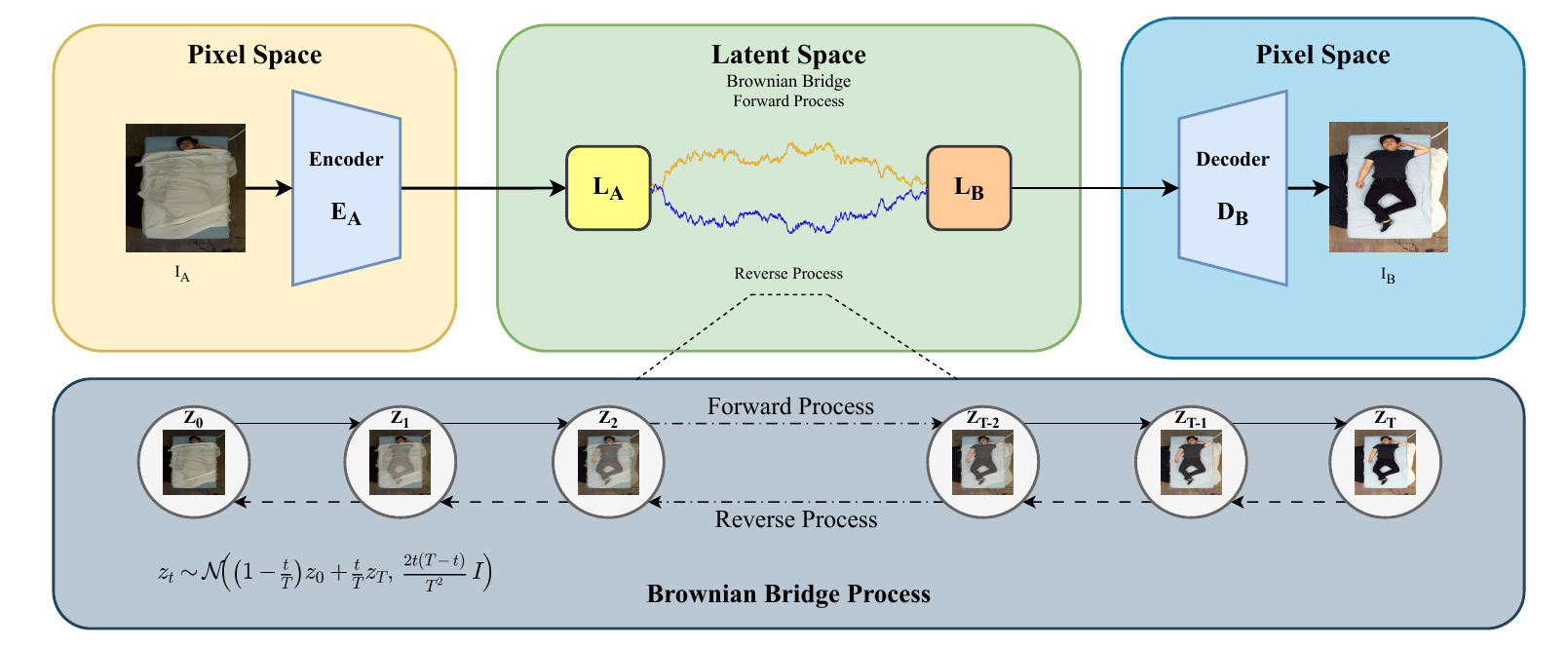}
\caption{
Brownian Bridge Diffusion Model (BBDM) for paired domain translation~\cite{li2023}. 
An uncovered image $U$ and its paired covered counterpart $C$ are encoded into latent representations $z_U$ and $z_C$. 
The lower panel illustrates the Brownian bridge distribution 
$z_t \sim \mathcal{N}((1-\tfrac{t}{T})z_U + \tfrac{t}{T}z_C,\; \tfrac{t(T-t)}{T}I)$, 
which explicitly connects the two endpoints in latent space. 
In the text, this bridge is expressed equivalently in parametric form as in Eq.~\eqref{eq:bbdm}. 
The reverse diffusion process predicts the injected noise to traverse the bridge from $z_U$ toward $z_C$, and the final latent state is decoded to produce the translated image.
}
    \label{fig:bbdm}
\end{figure}

\paragraph{Paired Translation (BBDM)}
When aligned uncovered–covered image pairs are available, we employ the Brownian Bridge Diffusion Model (BBDM)~\cite{li2023} as a representative paired image-to-image translation baseline (see Fig.~\ref{fig:bbdm}). Unlike standard unconditional diffusion models (e.g., DDPM~\cite{ho2020ddpm}), BBDM is explicitly designed for paired translation by modeling a stochastic Brownian Bridge process between the latent representations of a source image $U$ and its aligned target counterpart $C$. The uncovered image is first encoded into latent space, where the diffusion bridge connects the latent states of $U$ and $C$, and the decoded output produces the translated image.

Conceptually, the forward bridge process interpolates between the latent encodings $z_U$ and $z_C$ while injecting Gaussian noise:
\begin{equation}
z_t = (1 - m_t)z_U + m_t z_C + \sqrt{\delta_t}\,\epsilon,
\label{eq:bbdm}
\end{equation}
where $\epsilon \sim \mathcal{N}(0, I)$ and $m_t$, $\delta_t$ denote time-dependent interpolation and noise schedules. The model learns to reverse this process by predicting the injected noise residual:
\begin{equation}
\mathcal{L}_{BBDM} =
\mathbb{E}_{z_U, z_C, \epsilon, t}
\Big[
\| m_t (z_C - z_U)
+ \sqrt{\delta_t}\,\epsilon
- \epsilon_{\phi}(z_t, t) \|^2
\Big].
\end{equation}

During inference, covered samples are obtained by sampling the learned reverse bridge process conditioned on the encoded latent representation of the uncovered image $U$, followed by decoding to pixel space.

By explicitly modeling the transition between paired samples, BBDM enforces stronger geometric alignment than unpaired translation methods. However, similar to CycleGAN, the generation process remains conditioned on a visible source image, restricting synthesis to pose configurations present in the uncovered dataset. \\

\begin{figure}[tb!]
    \centering
    \includegraphics[width=\linewidth]{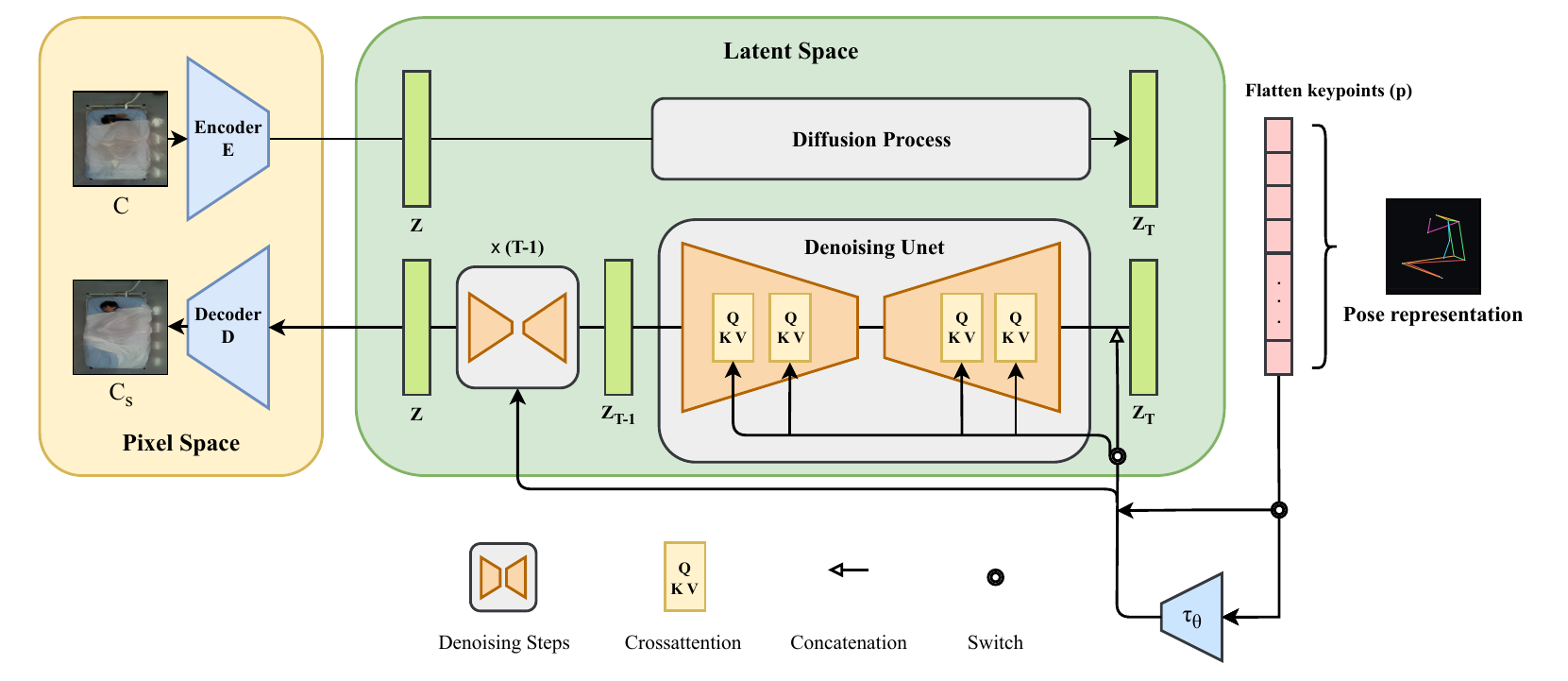}
\caption{
Pose-Conditioned Latent Diffusion Model (Pose-LDM). 
A covered image $C$ is encoded by $\mathcal{E}$ into a latent representation $z_0$. 
The forward diffusion process progressively corrupts $z_0$ into $z_t$ via Gaussian noise. 
The denoising U-Net $\epsilon_\theta$ \cite{ronneberger2015unet} predicts the injected noise while being conditioned on skeletal keypoints $p$ through a learnable embedding $\tau_\theta(p)$, injected via either spatial concatenation or cross-attention. 
During inference, generation begins from Gaussian noise $z_T \sim \mathcal{N}(0,I)$ and the learned reverse diffusion process produces a latent sample that is decoded by $\mathcal{D}$ to yield the synthesized covered image $C_s$.
}
    \label{fig:ldm}
\end{figure}

\subsubsection{Pose-Conditioned Latent Diffusion (Pose-LDM: Proposed Approach)}

To move beyond image-to-image translation, we propose a pose-conditioned generative framework based on a Conditional Latent Diffusion Model (LDM)~\cite{rombach2022high} (Fig.~\ref{fig:ldm}). We adopt LDM due to its ability to perform stable and high-fidelity image synthesis in a compressed latent space while supporting flexible conditioning mechanisms such as cross-attention~\cite{rombach2022high,podell2023sdxl}. Compared to pixel-space diffusion models~\cite{ho2020ddpm}, latent diffusion reduces computational cost while preserving generative quality, and enables effective integration of structured pose embeddings~\cite{cao2024survey}. Unlike image-guided methods, the proposed model synthesizes covered images directly from pose keypoints $p$, without conditioning on a visible source image.

\paragraph{Latent Diffusion Formulation}

Following~\cite{rombach2022high}, generation is performed in a lower-dimensional latent space. Let $\mathcal{E}(\cdot)$ and $\mathcal{D}(\cdot)$ denote the encoder and decoder of a variational autoencoder. A real covered image $C$ is first mapped to a latent representation: 
$z_0 = \mathcal{E}(C)$.

The forward diffusion process progressively corrupts this latent representation with Gaussian noise:
\begin{equation}
z_t = \sqrt{\alpha_t}\, z_0 + \sqrt{1-\alpha_t}\, \epsilon,
\end{equation}
where $\epsilon \sim \mathcal{N}(0,I)$ and $\alpha_t = \prod_{s=1}^{t} (1-\beta_s)$ follows the standard diffusion noise schedule~\cite{ho2020ddpm}.

The denoising U-Net $\epsilon_\theta$ is trained to predict the injected noise conditioned on pose:
\begin{equation}
\mathcal{L}_{LDM} =
\mathbb{E}_{C, p, \epsilon, t}
\left[
\| \epsilon - \epsilon_\theta(z_t, t, \tau_\theta(p)) \|_2^2
\right],
\label{eq:ldm_loss}
\end{equation}
where $\tau_\theta(p)$ denotes a learnable embedding that maps raw keypoint coordinates into the conditioning space.

After the reverse diffusion process, the predicted latent $\hat{z}_0$ is mapped back to image space via the decoder, yielding the synthesized covered image ${C_s} = \mathcal{D}(\hat{z}_0)$. 

During inference, generation begins from Gaussian noise $z_T \sim \mathcal{N}(0,I)$ and iteratively applies the learned reverse diffusion process conditioned on pose to obtain $\hat{z}_0$, which is then decoded to produce the synthesized image.

\paragraph{Pose Conditioning Mechanisms}

Crucially, the conditioning input consists strictly of raw keypoint coordinates rather than rendered skeleton images or heatmaps, preserving geometric abstraction and avoiding additional visual bias.

We investigate two integration mechanisms for injecting pose information into the denoising network:
\begin{itemize}[leftmargin=*, nosep]
    \item \textbf{Spatial Concatenation.} The pose representation is processed through an MLP ($\tau_\theta$), spatially aligned, and concatenated with the noisy latent feature map at the network input.
    \item \textbf{Cross-Attention.} The pose embeddings are projected via an MLP and injected as keys and values into intermediate cross-attention layers of the U-Net, enabling dynamic global interaction between skeletal geometry and latent features.
\end{itemize}

Unlike BBDM, which performs paired translation conditioned on a source image, Pose-LDM performs latent diffusion conditioned solely on skeletal embeddings, eliminating dependence on a visible source image.
This removes dependence on aligned uncovered–covered image pairs and enables synthesis from arbitrary pose inputs while keeping the downstream pose estimator fixed across all augmentation strategies.

This pose-conditioned formulation constitutes the main contribution of our work. By decoupling generation from visible source images, the model enables synthesis of covered samples from arbitrary pose inputs, increasing diversity of synthetic training data. Furthermore, because conditioning relies exclusively on abstract skeletal representations rather than raw appearance cues, the approach reduces direct conditioning on identifiable appearance features during generation in sensitive monitoring environments.

\subsection{Overall Procedure}
All augmentation strategies follow a unified three-stage protocol to ensure fair comparison. 
First, the generative model (CycleGAN, BBDM, or Pose-LDM) is trained on the training split to learn either an image-to-image translation (for CycleGAN and BBDM) or a pose-to-image synthesis mapping (for Pose-LDM). 
Second, the trained generator is used to synthesize a dataset of covered images $C_s$ from uncovered samples or their associated pose annotations. 
Because the augmentation is designed to preserve body geometry, ground-truth pose labels from the uncovered dataset remain valid for the synthesized images. 
Finally, a fixed pose estimation backbone is fine-tuned on the combined dataset $U + C_s$. 
All methods are evaluated primarily based on downstream pose estimation performance under blanket occlusion, isolating the impact of the generative augmentation strategy.

\section{Experimental Analysis}
\label{sec:experiments}

\subsection{Dataset}
We evaluate our proposed method on the Simultaneously-collected Multimodal Lying Pose (SLP) dataset~\cite{liu2020slp}. SLP is a large-scale benchmark specifically designed for in-bed human pose estimation and comprises data from 109 subjects. The samples are captured in two distinct physical settings, a home bedroom and a hospital room, using multiple sensing modalities, including RGB, Long Wave Infrared (LWIR), Depth, and Pressure maps. To ensure diversity and robustness, we utilize samples from both environments, capturing variations in background clutter, bed configurations, and lighting conditions. For each subject, images are recorded under three covering conditions: uncovered, covered by a thin sheet, and covered by a thick blanket. In our experiments, we rely exclusively on the RGB modality to simulate standard consumer-grade camera setups.

\subsection{Evaluation Metrics}
Performance is evaluated using two complementary metrics: PCK@0.1 and AP$_{50\text{–}95}$. 
PCK@0.1 measures the percentage of correctly localized keypoints under a strict localization threshold, emphasizing geometric precision under severe occlusion. 
AP$_{50\text{–}95}$ follows the standard COCO-style \cite{lin2014coco} evaluation protocol, averaging average precision across multiple IoU thresholds, and reflects overall detection robustness.

\subsection{Implementation Details}

All experiments were conducted on NVIDIA RTX 4090 GPUs (24 GB memory) using PyTorch. Unless otherwise specified, the reported hyperparameters correspond to the final configurations used for the results presented in Sec.~\ref{subsec:results}. All hyperparameters were selected based on validation performance, and experiments were run with a fixed random seed.

For the deterministic heuristic augmentation, padding parameters were sampled from $\delta_x, \delta_y \sim \mathcal{U}(a,b)$ with $a=20$ and $b=40$ pixels. The Gaussian smoothing kernel used for mask softening employed a random kernel size selected from the choices $\{7, 9, 11\}$ with standard deviation $\sigma \in [1.4, 2.0]$, creating soft fabric boundaries.

For the unpaired translation setting, CycleGAN~\cite{zhu2017} was implemented using a ResNet-based generator with 9 residual blocks and a PatchGAN discriminator. The model was optimized using Adam with a learning rate of $2 \times 10^{-4}$ for 200 epochs and a batch size of 8.

For the paired translation baseline, BBDM~\cite{li2023} employed a U-Net architecture with 128 base channels and attention resolutions at scales $\{32, 16, 8\}$. The diffusion process used $T=1000$ timesteps with a linear noise schedule and was trained for 200 epochs using Adam with a learning rate of $1 \times 10^{-4}$.

For the proposed Pose-LDM, we utilized a KL-regularized autoencoder with a U-Net backbone consisting of 192 base channels. The latent diffusion process was performed with $T=1000$ timesteps and a linear schedule. The autoencoder operated with a spatial downsampling factor of $f = 8$. The model was trained for 200 epochs with a base learning rate of $1 \times 10^{-4}$ using Adam.

Pose conditioning was implemented via a dedicated MLP embedder ($\text{Linear} \rightarrow \text{SiLU} \rightarrow \text{Linear}$). In the spatial concatenation setting, the embedding dimension was 128, whereas in the cross-attention setting it was 512 to match the transformer context. We evaluated two keypoint representations: 2D coordinates (input dimension 26) and coordinates augmented with visibility flags (input dimension 39).

For all generative models (CycleGAN, BBDM, and Pose-LDM), we adopted a subject-wise 70/15/15 split for training, validation, and testing, ensuring that identities in the test set were not observed during training. The same subject partition was maintained across all generative experiments to guarantee fair comparison.

For the downstream pose estimation task, we used the YOLOv11-Medium architecture~\cite{yolo11_ultralytics}, initialized with COCO-pretrained weights. YOLO-based keypoint detectors are widely adopted for real-time RGB pose estimation due to their favorable tradeoff between accuracy and computational efficiency \cite{dong2024mda,zhang2024sp,11020018}. Using a single, fixed backbone across all experiments allows us to isolate the effect of the generative augmentation strategy rather than improvements attributable to changes in the pose estimator itself.
The model was trained on a mixture of real uncovered images $U$ and synthetic covered images $C_s$ generated by the corresponding augmentation method under evaluation. A subject-wise 80/10/10 split was used for pose estimation training, validation, and testing, derived consistently from the same overall subject partition to avoid identity leakage. The pose estimator was trained for 50 epochs with a batch size of 64.

\subsection{Results}
\label{subsec:results}

We evaluate all augmentation strategies by fine-tuning a fixed YOLOv11-Medium pose estimator~\cite{yolo11_ultralytics} on a mixture of real uncovered images and synthetic covered images generated by each method. Performance (Table~\ref{tab:main_results}) is reported on the RGB modality of the SLP test set under two settings: (i) the full test set containing both covered and uncovered samples, and (ii) only covered samples. The latter setting isolates robustness under severe blanket occlusion.
For reference, we include two additional baselines. The \emph{Fully Supervised} setting denotes training the pose estimator on real labeled images from all covering conditions (uncovered, thin, and thick), representing an upper-bound scenario with access to annotated covered data. We also report the zero-shot performance of a pre-trained YOLO model and a model fine-tuned exclusively on real uncovered images without exposure to covered samples during training.

\begin{table*}[tb!]
  \centering
  \caption{
  Comparison of augmentation strategies on the RGB modality of the SLP test set. 
  Performance is measured using PCK@0.1 and AP$_{50\text{–}95}$. 
  No Paired Supervision: training does not require aligned uncovered–covered image pairs.
No Pixel Conditioning: generation does not use a visible uncovered RGB image as input.
Source-Free Inference: covered images can be synthesized without a source image at inference time.
Pose Expansion: method can generate samples from arbitrary or unseen pose configurations.
  }
  \label{tab:main_results}
  \resizebox{\textwidth}{!}{
  \begin{tabular}{lcccccccc}
    \toprule
    \multirow{2}{*}{\textbf{}} 
    & \multicolumn{4}{c}{\textbf{Augmentation Properties}} 
    & \multicolumn{4}{c}{\textbf{SLP Test Performance}} \\
    \cmidrule(lr){2-5} \cmidrule(lr){6-9}
    & No Paired & No Pixel & Source-Free & Pose 
    & \multicolumn{2}{c}{PCK@0.1} 
    & \multicolumn{2}{c}{AP$_{50\text{–}95}$} \\
    & Supervision & Conditioning & Inference & Expansion
    & Covered & Cov+Uncov
    & Covered & Cov+Uncov \\
    \midrule
    Pre-trained YOLO (zero-shot)
      & -- & -- & -- & -- 
      & 0.45 & 0.61 
      & 0.01 & 0.21 \\

    YOLO fine-tuned on Real Uncovered Data 
     & -- & -- & -- & -- 
      & 0.09 & 0.38 
      & 0.22 & 0.59 \\

    Fully Supervised (Real Uncovered + Real Covered)
      & -- & -- & -- & -- 
      & 0.97 & 0.98 
      & 0.90 & 0.92 \\
    \midrule
    Heuristic Augmentation + YOLO 
      & \cmark & \xmark & \xmark & \xmark
      & 0.50 & 0.67 
      & 0.32 & 0.63 \\

    CycleGAN Synthesis + YOLO
      & \cmark & \xmark & \xmark & \xmark
      & 0.80 & 0.86 
      & 0.55 & 0.73 \\

    BBDM Synthesis + YOLO
      & \xmark & \xmark & \xmark & \xmark
      & 0.91 & 0.94
      & \textbf{0.76} & \textbf{0.84} \\

    \rowcolor{lightblue}
    Pose-LDM (Ours) + YOLO
      & \cmark & \cmark & \cmark & \cmark
      & \textbf{0.93} & \textbf{0.95} 
      & 0.75 & 0.82 \\

    \bottomrule
  \end{tabular}}
\end{table*}

\begin{table*}[tb!]
\centering
\begin{threeparttable}
\caption{Performance comparison of YOLOv11 trained using synthetic data generated by different LDM conditioning strategies. Results are reported using PCK@0.1 and AP50-95.}
\label{tab:ldm_ablation}
\footnotesize
\renewcommand\arraystretch{1.2}

\begin{tabularx}{\textwidth}{l l l 
>{\centering\arraybackslash}X 
>{\centering\arraybackslash}X
>{\centering\arraybackslash}X 
>{\centering\arraybackslash}X}
\toprule
\multirow{2}{*}{\textbf{Conditioning Injection}} &
\multirow{2}{*}{\textbf{Pose Representation}} &
\multirow{2}{*}{\textbf{Encoding}} &
\multicolumn{2}{c}{\textbf{Covered + Uncovered}} &
\multicolumn{2}{c}{\textbf{Only Covered}} \\
\cmidrule(lr){4-5} \cmidrule(lr){6-7}
& & & PCK@0.1 & AP50--95 & PCK@0.1 & AP50--95 \\
\midrule

\multirow{4}{*}{Spatial Concatenation}
& 2D pose + visibility & Direct & 0.945 & 0.797 & 0.924 & 0.708 \\
& 2D pose + visibility & MLP    & 0.933 & 0.783 & 0.904 & 0.682 \\
& 2D pose              & Direct & 0.937 & 0.779 & 0.912 & 0.684 \\
& 2D pose              & MLP    & 0.918 & 0.753 & 0.883 & 0.624 \\

\midrule

\multirow{2}{*}{Cross-Attention}
& 2D pose + visibility & MLP & 0.948 & 0.817 & 0.928 & 0.744 \\
& 2D pose              & MLP & \textbf{0.950} & \textbf{0.819} & \textbf{0.930} & \textbf{0.745} \\

\bottomrule
\end{tabularx}
\end{threeparttable}
\end{table*}

\subsubsection{Ablation Study on Pose-LDM}

To analyze the impact of conditioning design within Pose-LDM, we conduct an ablation study over three components: (i) pose representation, (ii) pose encoding strategy, and (iii) conditioning injection mechanism.
For the pose representation, we evaluate two input variants. The first consists of raw 2D joint coordinates $(x,y)$ for each keypoint (denoted as ``2D pose''). The second augments these coordinates with a per-joint binary occlusion prior (denoted as ``2D pose + visibility''). This indicator is heuristically defined: for covered images, joints anatomically below the head are marked as occluded (0), while visible joints are marked as 1.
For the pose encoding strategy, we compare two approaches for mapping pose inputs into the conditioning space of the diffusion model. In the \emph{Direct} setting, the pose vector is reshaped and aligned to match the latent spatial resolution without additional learnable transformation. In the \emph{MLP} setting, the pose vector is projected through a lightweight multi-layer perceptron before injection, enabling learned feature adaptation prior to conditioning.
Finally, we compare two conditioning injection mechanisms for integrating pose information into the denoising U-Net. In \emph{Spatial Concatenation}, the encoded pose representation is spatially broadcast and concatenated with latent feature maps at the network input. In \emph{Cross-Attention}, pose embeddings are injected as conditioning tokens within intermediate attention layers, allowing dynamic global interaction between pose features and latent representations during denoising.

Table~\ref{tab:ldm_ablation} reports the performance of the aforementioned conditioning configurations within Pose-LDM. 
The best overall performance is achieved using Cross-Attention with raw 2D pose coordinates and MLP encoding.
Comparing fusion strategies, Cross-Attention consistently outperforms Spatial Concatenation across all pose representations and metrics. 
The improvement is observed for both metrics, indicating that dynamically integrating pose information through attention is more effective than static spatial concatenation.
Regarding the pose encoding strategy, under Spatial Concatenation, MLP encoding generally improves performance compared to Direct reshaping, suggesting that projecting pose coordinates into a learned feature space facilitates better alignment with latent representations. 
Under Cross-Attention, the performance gap between encoding strategies is smaller, indicating that attention-based conditioning is less sensitive to the specific encoding formulation. Finally, augmenting coordinates with the heuristic occlusion prior does not provide consistent improvements. In fact, the strongest configuration relies solely on raw 2D coordinates, suggesting that the diffusion model can infer occlusion structure directly from pose geometry without requiring explicit occlusion indicators. Overall, these results indicate that the conditioning injection mechanism plays a more significant role than the specific pose representation, and that geometry-only conditioning is sufficient to achieve competitive occlusion-aware augmentation.

\subsubsection{Qualitative Analysis}

\begin{figure*}[tb!]
    \centering
    \includegraphics[width=0.53\textwidth]{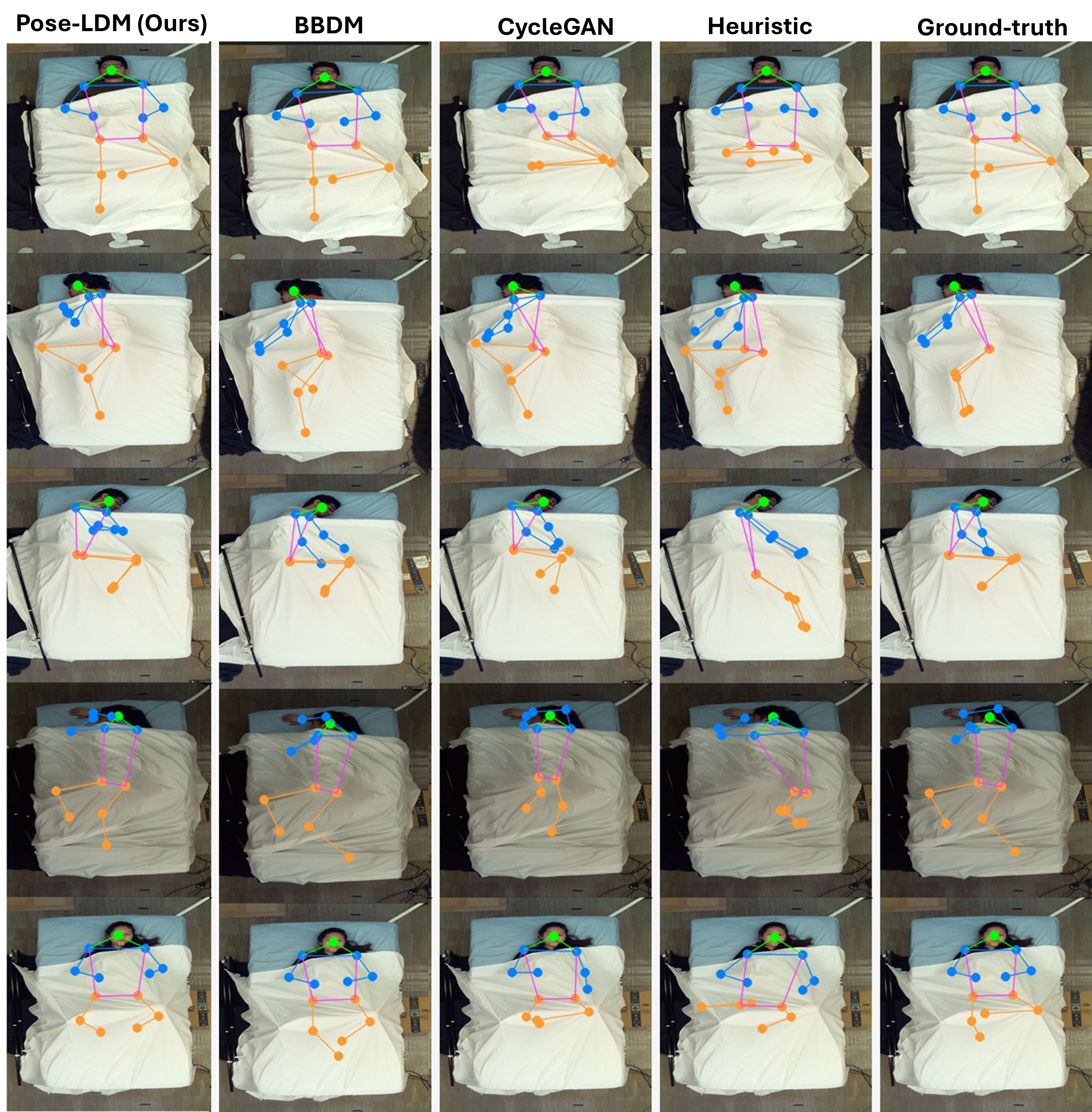}
    \caption{
    Qualitative comparison of pose estimation results under blanket occlusion. 
    Each column corresponds to a pose estimator trained using a different augmentation strategy: Pose-LDM (ours), BBDM, CycleGAN, heuristic masking, and ground truth (GT). 
    Diffusion-based augmentation methods produce more stable and anatomically consistent joint predictions under severe occlusion. In particular, Pose-LDM preserves lower-body joint structure more reliably than heuristic and unpaired translation approaches, while achieving comparable structural alignment to paired diffusion without requiring paired supervision or pixel-level source conditioning.
    }
    \label{fig:qualitatives}
\end{figure*}

Qualitative comparisons in Fig.~\ref{fig:qualitatives} further illustrate that diffusion-based augmentation leads to more stable and anatomically consistent joint predictions under severe blanket occlusion. Compared to heuristic masking and unpaired translation, Pose-LDM produces predictions that more closely align with ground-truth geometry, particularly for heavily occluded lower-body joints. While paired diffusion (BBDM) exhibits similarly strong structural preservation, Pose-LDM achieves comparable stability without relying on paired supervision or pixel-level source conditioning.

It is important to note that the objective of this work is not high-fidelity blanket synthesis, realistic texture modeling, physically plausible fold simulation, or visual indistinguishability from real images. Nevertheless, for interested readers, Pose-LDM-generated images are provided in the Supplementary Material.

\section{Conclusion}
\label{sec:conc}

In this work, we presented a geometry-conditioned approach to occlusion-aware in-bed pose augmentation. Rather than translating visible images into covered ones, we introduced a pose-conditioned Latent Diffusion Model (Pose-LDM) that synthesizes blanket occlusions directly from skeletal keypoints. This design removes the need for paired uncovered-covered supervision and eliminates reliance on pixel-level source-image conditioning.

Through a systematic comparison with heuristic masking, unpaired translation, and paired diffusion-based translation, we demonstrated that geometry-conditioned diffusion achieves the strongest joint localization performance under severe occlusion while remaining competitive in overall detection accuracy. Notably, the best-performing configuration relies solely on raw 2D pose coordinates, indicating that explicit occlusion priors are unnecessary and that pose geometry alone provides sufficient structural information for effective conditioning.
Our ablation study further showed that the conditioning injection mechanism plays a decisive role, with cross-attention consistently outperforming spatial concatenation. This highlights the importance of dynamic interaction between pose embeddings and latent features in guiding occlusion synthesis.

The proposed framework is evaluated on SLP~\cite{liu2020slp}, currently the only large-scale benchmark for in-bed pose estimation under realistic blanket occlusion. In the absence of other comparable public datasets, this evaluation represents the strongest available testbed for the considered task. Future work will explore extending the approach to additional sensing modalities, more diverse environments, and richer structural representations such as 3D pose or parametric body models.



\section{Supplementary Material}

\author{\parbox{16cm}{\centering
    {\large Navid Aslankhani Khameneh$^{1}$, Marco Carletti$^2$, and Cigdem Beyan$^1$}\\
    {\normalsize
    $^1$Department of Computer Science, University of Verona, Verona, Italy \\
    $^2$EVS - Embedded Vision Systems Srl, Verona, Italy}}
}

\begin{figure*}[tb!]
    \centering

    \includegraphics[width=0.95\textwidth]{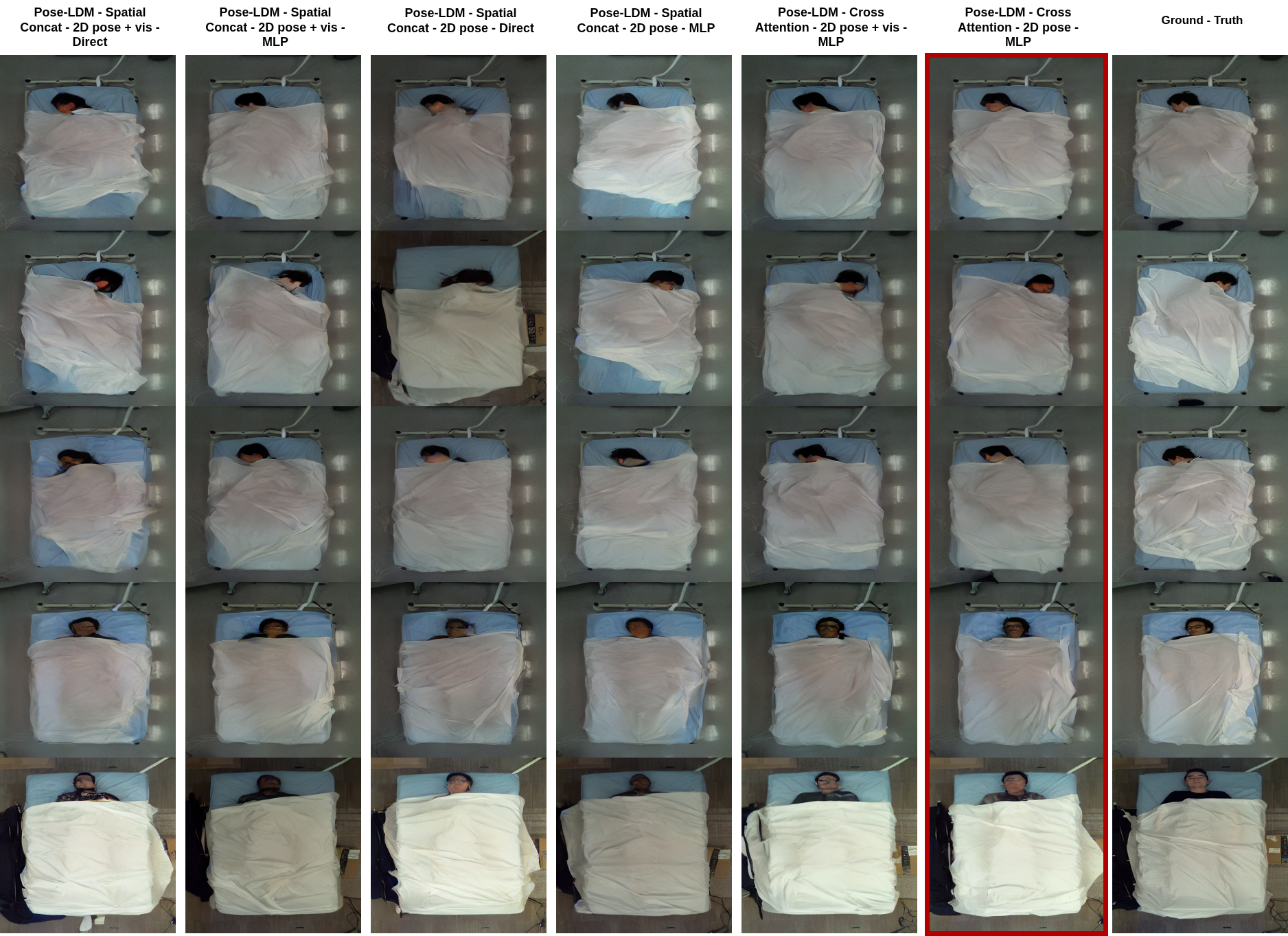}
    
    \vspace{5mm}

    \includegraphics[width=0.95\textwidth]{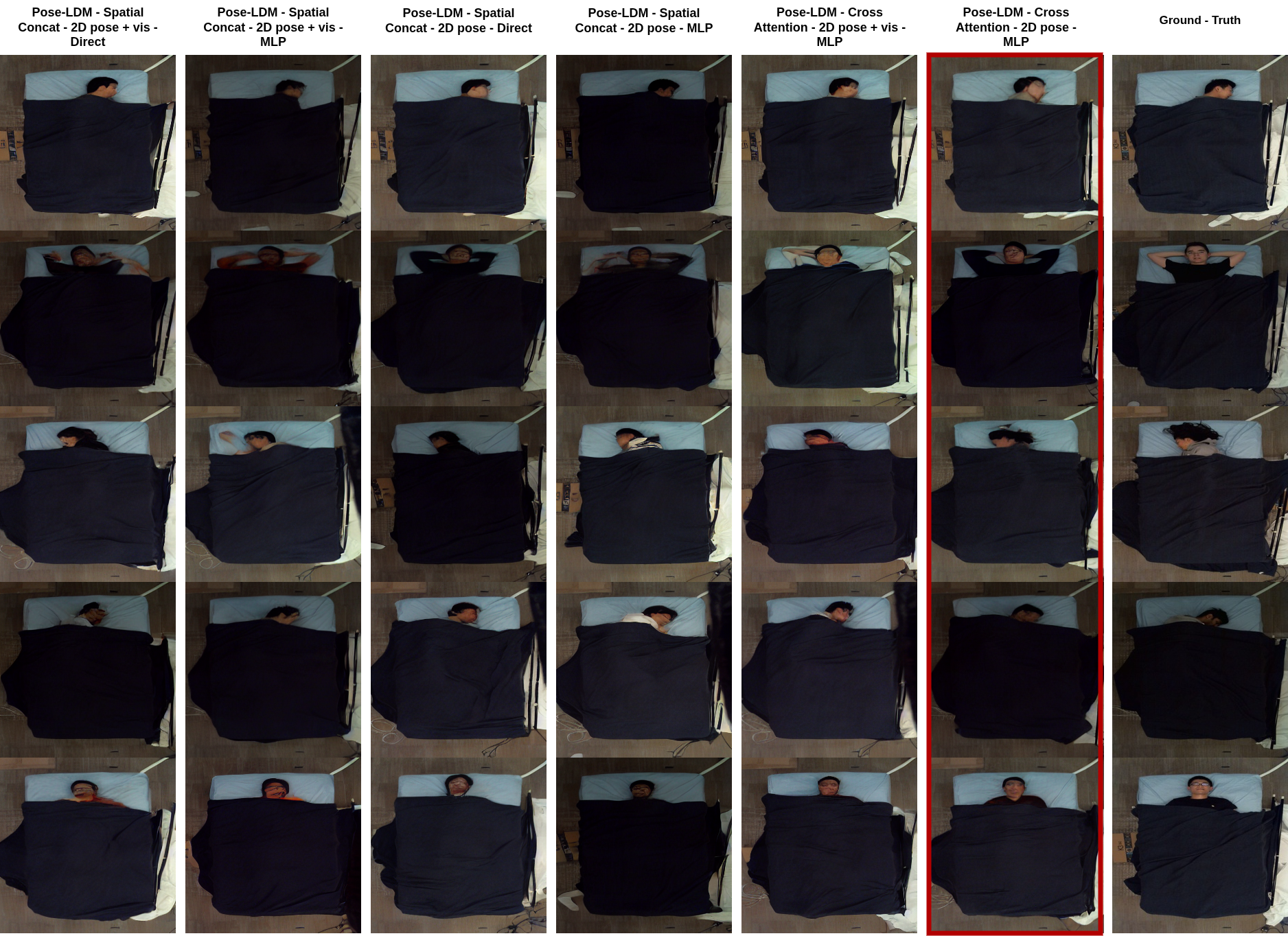}
    
    \caption{Examples of samples generated under covered settings using different conditioning variants of our Pose-LDM with two different bedsheet types.}
    \label{fig:bedsheet_types}
\end{figure*}

\subsection{Additional Qualitative Results}

Fig.~\ref{fig:bedsheet_types} presents qualitative samples generated under covered settings using different conditioning variants of Pose-LDM. Despite variations in surface texture, lighting, and shading, the generated outputs remain visually coherent and anatomically plausible, maintaining consistent global body structure relative to the conditioned poses.

These examples complement the quantitative results reported in Table~2 of the main manuscript by illustrating how different conditioning strategies influence structural consistency and occlusion modeling. Configurations with higher PCK and AP scores in Table~2 tend to produce visually more stable body structures and fewer apparent artifacts in the generated samples. Overall, the qualitative results provide complementary visual insight into the performance trends observed in the quantitative evaluation.

{\small
\bibliographystyle{ieee}
\bibliography{egbib}

@inproceedings{zhu2017,
  title={Unpaired Image-to-Image Translation using Cycle-Consistent Adversarial Networks},
  author={Zhu, Jun-Yan and Park, Taesung and Isola, Phillip and Efros, Alexei A},
  booktitle={Proceedings of the IEEE International Conference on Computer Vision (ICCV)},
  pages={2223--2232},
  year={2017}
}

@inproceedings{li2023,
  title={BBDM: Image-to-Image Translation with Brownian Bridge Diffusion Models},
  author={Li, Bo and Xue, Kaitao and Liu, Bin and Lai, Yu-Kun},
  booktitle={Proceedings of the IEEE/CVF Conference on Computer Vision and Pattern Recognition (CVPR)},
  pages={1952--1961},
  year={2023}
}

@incollection{loper2023smpl,
  title={SMPL: A skinned multi-person linear model},
  author={Loper, Matthew and Mahmood, Naureen and Romero, Javier and Pons-Moll, Gerard and Black, Michael J},
  booktitle={Seminal Graphics Papers: Pushing the Boundaries, Volume 2},
  pages={851--866},
  year={2023}
}

@inproceedings{tarvainen2017mean,
  title={Mean teachers are better role models: Weight-averaged consistency targets improve semi-supervised deep learning results},
  author={Tarvainen, Antti and Valpola, Harri},
  booktitle={Advances in Neural Information Processing Systems},
  volume={30},
  year={2017}
}

@article{cao2019openpose,
  title={Openpose: Realtime multi-person 2d pose estimation using part affinity fields},
  author={Cao, Zhe and Hidalgo, Gines and Simon, Tomas and Wei, Shih-En and Sheikh, Yaser},
  journal={IEEE transactions on pattern analysis and machine intelligence},
  volume={43},
  number={1},
  pages={172--186},
  year={2019},
  publisher={IEEE}
}

@inproceedings{ho2020ddpm,
  title={Denoising Diffusion Probabilistic Models},
  author={Ho, Jonathan and Jain, Ajay and Abbeel, Pieter},
  booktitle={Advances in Neural Information Processing Systems},
  volume={33},
  pages={6840--6851},
  year={2020}
}

@article{pang2021image,
  title={Image-to-image translation: Methods and applications},
  author={Pang, Yingxue and Lin, Jianxin and Qin, Tao and Chen, Zhibo},
  journal={IEEE Transactions on Multimedia},
  volume={24},
  pages={3859--3881},
  year={2021},
  publisher={IEEE}
}

@inproceedings{nyamathulla2024analysis,
  title={Analysis of pix2pix and cyclegan for image-to-image translation: A comparative study},
  author={Nyamathulla, S and Veeranjaneyulu, N},
  booktitle={2024 IEEE International Conference on Smart Power Control and Renewable Energy (ICSPCRE)},
  pages={1--6},
  year={2024},
  organization={IEEE}
}

@article{dou2020asymmetric,
  title={Asymmetric CycleGAN for image-to-image translations with uneven complexities},
  author={Dou, Hao and Chen, Chen and Hu, Xiyuan and Jia, Libang and Peng, Silong},
  journal={Neurocomputing},
  volume={415},
  pages={114--122},
  year={2020},
  publisher={Elsevier}
}

@inproceedings{torbunov2023uvcgan,
  title={Uvcgan: Unet vision transformer cycle-consistent gan for unpaired image-to-image translation},
  author={Torbunov, Dmitrii and Huang, Yi and Yu, Haiwang and Huang, Jin and Yoo, Shinjae and Lin, Meifeng and Viren, Brett and Ren, Yihui},
  booktitle={Proceedings of the IEEE/CVF winter conference on applications of computer vision},
  pages={702--712},
  year={2023}
}

@inproceedings{newell2016stacked,
  title={Stacked hourglass networks for human pose estimation},
  author={Newell, Alejandro and Yang, Kaiyu and Deng, Jia},
  booktitle={European conference on computer vision},
  pages={483--499},
  year={2016},
  organization={Springer}
}

@article{liu2020slp,
  title={Simultaneously-Collected Multimodal Lying Pose Dataset: Towards In-Bed Human Pose Monitoring Under Various Cover Conditions},
  author={Liu, Shuangjun and Ostadabbas, Sarah},
  journal={IEEE Transactions on Pattern Analysis and Machine Intelligence},
  volume={45},
  number={1},
  pages={536--550},
  year={2022},
  publisher={IEEE}
}

@inproceedings{maji2022yolopose,
  title={YOLO-Pose: Enhancing YOLO for Multi Person Pose Estimation using Object Keypoint Similarity Loss},
  author={Maji, Debapriya and Nagori, Soyeb and Mathew, Manu and Poddar, Deepak},
  booktitle={Proceedings of the IEEE/CVF Conference on Computer Vision and Pattern Recognition (CVPR) Workshops},
  pages={2637--2646},
  year={2022}
}

@misc{yolo11_ultralytics,
  title={Ultralytics YOLO11},
  author={Jocher, Glenn and Qiu, Jing},
  year={2024},
  publisher={GitHub},
  journal={GitHub repository},
  howpublished={\url{https://github.com/ultralytics/ultralytics}}
}

@inproceedings{afham2022,
  title={Towards Accurate Cross-Domain In-Bed Human Pose Estimation},
  author={Afham, Mohamed and Haputhanthri, Udith and Pradeepkumar, Jathushan and Anandakumar, Mithunjha and De Silva, Amal and Edussooriya, Chamira U S},
  booktitle={ICASSP 2022 - 2022 IEEE International Conference on Acoustics, Speech and Signal Processing (ICASSP)},
  pages={1--5},
  year={2022},
  organization={IEEE}
}

@inproceedings{yin2020,
  title={Multimodal In-Bed Pose and Shape Estimation Under the Blankets},
  author={Yin, Yu and Robinson, Joseph P and Fu, Yun},
  booktitle={Proceedings of the 28th ACM International Conference on Multimedia},
  pages={1--9},
  year={2020}
}

@article{dayarathna2022,
  title={Privacy-Preserving In-Bed Pose Monitoring: A Fusion and Reconstruction Study},
  author={Dayarathna, Thanuja and Muthukumarana, Thilina and Rathnayaka, Yasith and Denman, Simon and de Silva, Chamira and Pemasiri, Asanka and Ahmedt-Aristizabal, David},
  journal={Expert Systems with Applications},
  volume={203},
  pages={119139},
  year={2022},
  publisher={Elsevier}
}

@INPROCEEDINGS{cao2021,
  author={Cao, Ting and Armin, Mohammad Ali and Denman, Simon and Petersson, Lars and Ahmedt-Aristizabal, David},
  booktitle={2022 IEEE 19th International Symposium on Biomedical Imaging (ISBI)}, 
  title={In-Bed Human Pose Estimation from Unseen and Privacy-Preserving Image Domains}, 
  year={2022},
  volume={},
  number={},
  pages={1-5},
  doi={10.1109/ISBI52829.2022.9761598}}

@article{bigalke2022,
  title={Anatomy-Guided Domain Adaptation for 3D In-Bed Human Pose Estimation},
  author={Bigalke, Alexander and Hansen, Lasse and Diesel, Jasper and Hennigs, Carsten and Rostalski, Philipp and Heinrich, Mattias P},
  journal={Medical Image Analysis},
  volume={83},
  pages={102687},
  year={2023},
  publisher={Elsevier}
}

@article{davoodnia2021,
  title={Estimating Pose from Pressure Data for Smart Beds with Deep Image-Based Pose Estimators},
  author={Davoodnia, V and Ghorbani, S and Ostadabbas, S},
  journal={Applied Intelligence},
  volume={52},
  pages={3864--3877},
  year={2022},
  publisher={Springer}
}

@inproceedings{obeidavi2022pose,
  title={In-pose estimation of covered and uncovered human body from thermal camera images using Multi-Scale Stacked Hourglass (MSSHg) Network},
  author={Obeidavi, Sahereh and Gandomkar, Mojtaba and Hirtz, Gangolf},
  booktitle={2022 16th International Conference on Signal-Image Technology \& Internet-Based Systems (SITIS)},
  pages={84--90},
  year={2022},
  organization={IEEE}
}

@inproceedings{lin2014coco,
  title={Microsoft COCO: Common Objects in Context},
  author={Lin, Tsung-Yi and Maire, Michael and Belongie, Serge and et al.},
  booktitle={ECCV},
  year={2014}
}

@inproceedings{ronneberger2015unet,
  title={U-Net: Convolutional Networks for Biomedical Image Segmentation},
  author={Ronneberger, Olaf and Fischer, Philipp and Brox, Thomas},
  booktitle={MICCAI},
  year={2015}
}

@article{cao2024survey,
  title={A survey on generative diffusion models},
  author={Cao, Hanqun and Tan, Cheng and Gao, Zhangyang and Xu, Yilun and Chen, Guangyong and Heng, Pheng-Ann and Li, Stan Z},
  journal={IEEE transactions on knowledge and data engineering},
  volume={36},
  number={7},
  pages={2814--2830},
  year={2024},
  publisher={IEEE}
}

@article{podell2023sdxl,
  title={Sdxl: Improving latent diffusion models for high-resolution image synthesis},
  author={Podell, Dustin and English, Zion and Lacey, Kyle and Blattmann, Andreas and Dockhorn, Tim and M{\"u}ller, Jonas and Penna, Joe and Rombach, Robin},
  journal={arXiv preprint arXiv:2307.01952},
  year={2023}
}

@inproceedings{rombach2022high,
  title={High-Resolution Image Synthesis with Latent Diffusion Models},
  author={Rombach, Robin and Blattmann, Andreas and Lorenz, Dominik and Esser, Patrick and Ommer, Bj{\"o}rn},
  booktitle={Proceedings of the IEEE/CVF Conference on Computer Vision and Pattern Recognition (CVPR)},
  pages={10684--10695},
  year={2022}
}

@article{dong2024mda,
  title={MDA-YOLO Person: a 2D human pose estimation model based on YOLO detection framework},
  author={Dong, Chengang and Tang, Yuhao and Zhang, Liyan},
  journal={Cluster Computing},
  volume={27},
  number={9},
  pages={12323--12340},
  year={2024},
  publisher={Springer}
}

@article{zhang2024sp,
  title={SP-YOLO: An end-to-end lightweight network for real-time human pose estimation},
  author={Zhang, Yuting and Wang, Zongyan and Li, Menglong and Gao, Pei},
  journal={Signal, Image and Video Processing},
  volume={18},
  number={1},
  pages={863--876},
  year={2024},
  publisher={Springer}
}

@INPROCEEDINGS{11020018,
  author={Sonawane, Achal and Dandam, Varshini and Khamkar, Kiran and Wawge, Tanushri and More, Priyanka},
  booktitle={2025 International Conference on Computing and Communication Technologies (ICCCT)}, 
  title={Leveraging YOLO for Real-Time Human Detection and Pose Estimation in Live Stream Environments}, 
  year={2025},
  volume={},
  number={},
  pages={1-5},
  doi={10.1109/ICCCT63501.2025.11020018}}
}

\end{document}